\documentclass[conference]{IEEEtran}
\IEEEoverridecommandlockouts
\usepackage{cite}
\usepackage{amsmath,amssymb,amsfonts}
\usepackage{algorithmic}
\usepackage{graphicx}
\usepackage{textcomp}
\usepackage{xcolor}
\usepackage{algorithm}
\usepackage{multicol}
\usepackage{multirow}

\def\BibTeX{{\rm B\kern-.05em{\sc i\kern-.025em b}\kern-.08em
    T\kern-.1667em\lower.7ex\hbox{E}\kern-.125emX}}
\begin{document}

\title{MMDR: A Result Feature Fusion Object Detection Approach for Autonomous System
}

\author{\IEEEauthorblockN{1\textsuperscript{st} Wendong Zhang}
\IEEEauthorblockA{\textit{Department of Control Science and Engineering} \\
\textit{Harbin Institute of Technology}\\
Harbin, China \\
zavieton@163.com}
}
 
\maketitle

\begin{abstract}
Object detection has been extensively utilized in autonomous systems in recent years, encompassing both 2D and 3D object detection. 
Recent research in this field has primarily centered around multimodal approaches for addressing this issue.
In this paper, a multimodal fusion approach based on result feature-level fusion is proposed. This method utilizes the outcome features generated from single modality sources, and fuses them for downstream tasks.
Based on this method, a new post-fusing network is proposed for multimodal object detection, which leverages the single modality outcomes as features. The proposed approach, called Multi-Modal Detector based on Result features (MMDR), is designed to work for both 2D and 3D object detection tasks. 
Compared to previous multimodal models, the proposed approach in this paper performs feature fusion at a later stage, enabling better representation of the deep-level features of single modality sources. 
Additionally, the MMDR model incorporates shallow global features during the feature fusion stage, endowing the model with the ability to perceive background information and the overall input, thereby avoiding issues such as missed detections.
\end{abstract}


\section{Introduction}
In light of recent advancements in autonomous systems, the detection of two-dimensional and three-dimensional objects has garnered increasing attention from researchers. The ability to perceive surrounding objects has become increasingly critical for ensuring the safety of autonomous driving and enabling informed decision-making.
Over the past few years, the majority of research efforts in this area have focused on enhancing the performance and accuracy of 2D and 3D object detection methods, especially multimodal methods. 

There are two main methods for multimodal detection: decision-level fusion and feature-level fusion.
Feature-level fusion combines features extracted from different modalities to improve object detection performance, while decision-level fusion combines decisions made by different modalities to make the final decision for object detection.

In this paper, a result feature-level fusion method is proposed. In this method, a novel ``result feature" is proposed, which is constructed based on the results of single modality object detection and can be used for deeper feature fusion as well as for the subsequent detection tasks.

A multimodal model MMDR (Multi-Modal Detector based on Result features) based on the proposed result feature level fusion method is proposed in this paper.
In the MMDR model, deeper ``fusion features" are constructed based on the ``result feature". Meanwhile, some shallow ``global features" from single modality detection (first-stage detection) are preserved and are fused with the result feature to obtain the final ``fusion features". These features can express overall point cloud or image information and background information. Finally, for different detection tasks, the MMDR model performs further feature calculation on the ``fusion features" to obtain more accurate detection results (second-stage detection).

Our contribution can be summarized as following.
\begin{itemize}
    \item A novel method for multimodal detection that combines result features and global features to construct fusion features is proposed. This method involves preserving global features from single modality detection and fusing them with the result feature to obtain the final fusion features.
    \item A multimodal detection model called MMDR is proposed in this paper. This model based on the proposed result feature level fusion method, performs further feature calculation on the fusion features to obtain more accurate detection results.
    \item Experiments are conducted on KITTI dataset and the Apollo autonomous system to verify the performance of our model and the feasibility on real environments.
\end{itemize}

\section{Related Works}

\subsection{Three-dimensional detection method}

\subsubsection{Decision-level fusion method}
Decision-level fusion combines the decisions made by different modalities to make the final decision for object detection. For example, in a multimodal object detection system that uses both images and lidar data, the decisions made by each modality can be combined at a decision-level to make the final object detection decision. Decision-level fusion can be achieved through techniques such as majority voting, weighted voting, or stacking. 

There have been studies employing the decision-level fusion method. Qi \textit{et al.} proposed Frustum PointNet \cite{qi2018frustum} which is building upon the backbone network PointNet \cite{qi2017pointnet, qi2017pointnet++}. Frustum PointNet is a 3D object detection framework that utilizes both 2D and 3D information to enhance detection accuracy by addressing the challenge of detecting partially visible objects. The key innovation of this method is its approach to extracting 3D frustums from RGB-D images and applying PointNet for 3D object detection. Pang \textit{et al.} proposed CLOCs (Camera-LiDAR Object Candidates Fusion for 3D Object Detection)\cite{pang2020clocs} that fuses object candidates from both camera and LiDAR data to improve detection accuracy. This method uses candidate fusion to generate high-quality 3D bounding boxes, and achieves state-of-the-art performance on the KITTI benchmark.

\subsubsection{Feature-level fusion method}
Feature-level fusion combines the features extracted from different modalities to improve the performance of object detection. Feature-level fusion can be achieved through techniques such as concatenation, element-wise addition or multiplication, or attention-based methods that selectively weight and combine the features from different modalities. 
MV3D\cite{chen2017multi} is one of the foundational work by this method. It is a multimodal object detection framework that uses both LiDAR and camera data to improve detection accuracy, leveraging the complementary strengths of each sensor modality.
AVOD \cite{ku2018joint} propsed a multi-view based 3D object detection framework that generates 3D object proposals and performs object detection by aggregating features from multiple camera views.
Liang \textit{et al.} proposed ContFuse \cite{liang2018deep} which utilizes a continuous convolutional neural network to effectively fuse 3D LiDAR and RGB data, as well as its ability to handle missing or occluded data.
Vora \textit{et al.} proposed PointPainting\cite{vora2020pointpainting} to fuse point cloud and image segment results at the feature-level stage. This method projects high-dimensional features onto a 2D plane using a learned projection matrix, thereby leveraging the texture information of the input image to improve detection accuracy.

\begin{figure*}[htbp]
\centerline{\includegraphics[width = 17cm]{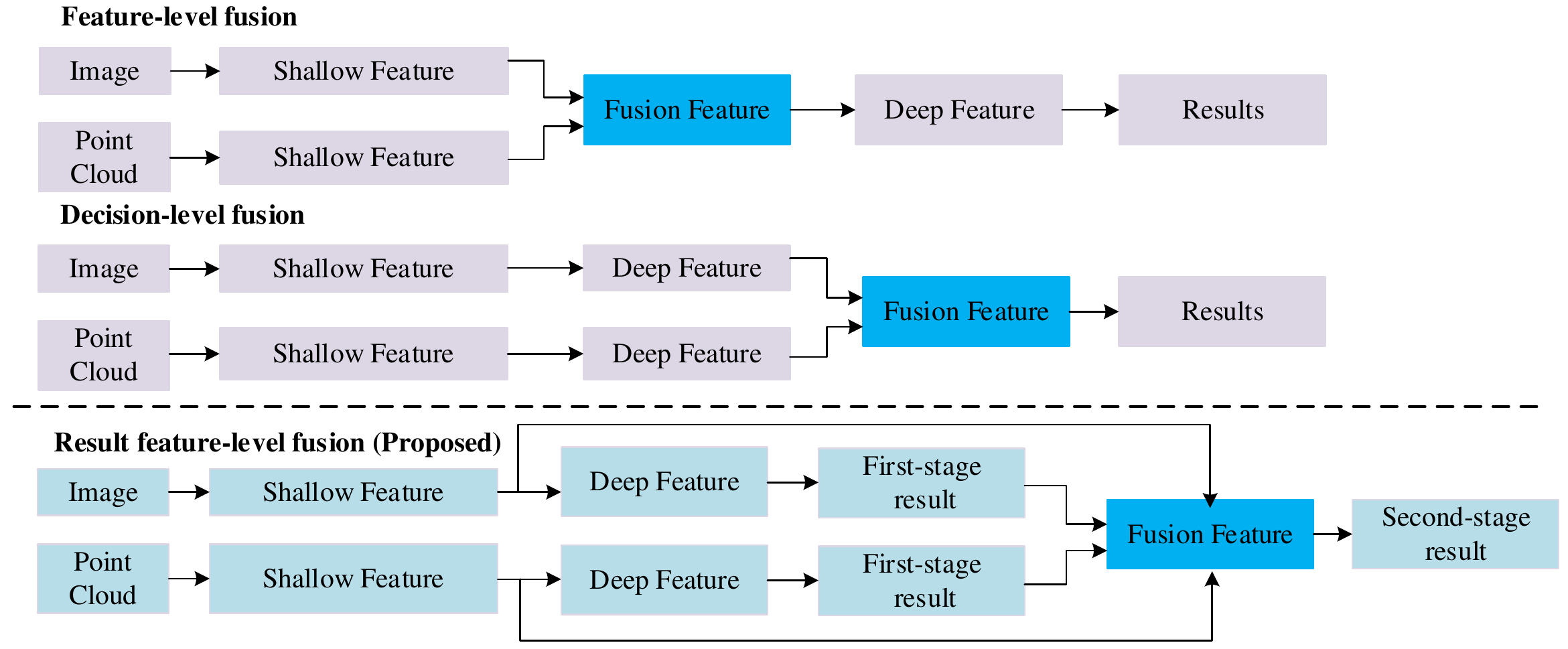}}
\caption{The comparison of three multimodal fusion methods: feature-level fusion, decision fusion and result feature-level fusion which is proposed in this paper.}
\label{fig2}
\end{figure*}

\subsubsection{Single modality method}
There have been several approaches proposed for estimating 3D bounding boxes from a single modality. For instance, Voxelnet \cite{zhou2018voxelnet} and SECOND \cite{yan2018second} divide a point cloud into 3D voxels and detect objects based on these voxels. PointNet \cite{qi2017pointnet, qi2017pointnet++} extracts features directly from point clouds. Other algorithms, such as PIXOR \cite{yang2018pixor} and PointPillars \cite{lang2019pointpillars}, extract features based on bird's-eye view (BEV). Additionally, Point-RCNN \cite{shi2019pointrcnn}, Point-GNN \cite{shi2020point}, PV-RCNN \cite{shi2020pv}, Local-Global Transformer \cite{pan20213d}, SE-SSD \cite{zheng2021se}, and CIA-SSD \cite{zheng2021cia} are proposed to learn features directly from the original point clouds.

\subsection{Two-dimensional detection method}
The current state-of-the-art object detection methods can be broadly categorized into two main trends. The first trend focuses on balancing real-time performance and accuracy using one-stage detectors, such as RetinaNet \cite{lin2017focal}, CenterNet \cite{duan2019centernet}, and YOLO series algorithms \cite{bochkovskiy2020yolov4, ge2021yolox, xu2022pp, huang2021pp, wang2021you, li2022yolov6, wang2022yolov7}.
Compared with other algorithms, they are generally faster and more computationally efficient as they do not require proposal generation and refinement steps. Therefore, they are well-suited for real-time applications and devices with limited computing power.

On the other hand, some methods concentrate on improving accuracy, especially in recent years with the popularity of the Transformer\cite{vaswani2017attention}. Vision Transformer \cite{dosovitskiy2020image} was the first to apply the Transformer architecture to computer vision tasks. Swin Transformer \cite{liu2021swin} computes with shifted windows and brings greater efficiency by limiting self-attention computation to non-overlapping local windows while also allowing for cross-window connection. Based on these studies, DETR \cite{carion2020end}, DINO \cite{zhang2022dino}, Swin Transformer v2 \cite{liu2022swin}, and Focal Transformer \cite{yang2021focal} have been proposed to solve the detection task and achieve better accuracy performance on benchmarks.

Recent research of two-dimensional object detection for autonomous system \cite{chen2021deep} has focused on scenes with only image inputs. However, it may be necessary to optimize algorithms using other modality information in some scenes.

\section{Architecture}

\subsection{Result feature-level fusion}
This paper proposes a result feature-level fusion method for multimodal 3D and 2D object detection. Compared to previous multimodal fusion methods at both feature-level and decision-level, this method achieves deeper feature fusion and stronger expression capabilities. 

Additionally, to improve the expression ability of background information and reduce missed and false detections resulting from first-stage detection, a "global feature" is integrated in the feature fusion stage. 

For the general detection network, the feature extraction network can be divided into three parts. Part A of the feature extraction network is used to extract shallow features from the input data, while Part B of the feature network is used to further extract features from the shallow features to obtain deeper features. Part C of the feature extraction network obtain the model output results from the deep features.

For the process of multimodal network, let $P$ denote the input point cloud and $I$ denote the input image. The inference process for feature-level fusion can be expressed as Eq.~\eqref{eq1}, and the decision feature-level fusion can be expressed as Eq.~\eqref{eq2}. The proposed result feature-level fusion can be expressed as Eq.~\eqref{eq3}.

\begin{equation}
\label{eq1}
\begin{cases}
f = A_{P}(P)+A_{I}(I) \\
r = C(B(f))
\end{cases}
\end{equation}
\begin{equation}
\label{eq2}
\begin{cases}
f = B_{P}(A_{P}(P)) + B_{I}(A_{I}(I)) \\
r = C(f)
\end{cases}
\end{equation}
\begin{equation}
\label{eq3}
\begin{cases}
f = rf + of \\
rf = C_{P}(B_{P}(A_{P}(P))) + C_{I}(B_{I}(A_{I}(I))) \\
of = D_{P}((A_{P}(P)) + D_{I}((A_{I}(I)) \\
r = E(f)
\end{cases}
\end{equation}

In these equations, $A_{P}$, $B_{P}$, and $C_{P}$ represent the corresponding network parts for processing point cloud data, while $A_{I}$, $B_{I}$, and $C_{I}$ represent the corresponding network parts for processing image data. $r$ denotes the model output result, $f$ represents the fused feature, $rf$ represents result feature, $of$ represents global feature, $D$ represents the network part for processing the global feature, and $E$ represents the network part for post-processing the fused feature in the proposed feature-level fusion approach in this paper.

The implementation comparison of various multimodal fusion methods is shown in Fig.~\ref{fig2}. In the result feature-level fusion of this figure, the first-stage result is the detection results obtained by a single modality detector, while the second-stage result corresponds to the final detection results based on the fusion feature.

\subsection{Process of MMDR model}
\begin{figure*}[htbp]
\centerline{\includegraphics[width = 17.5cm]{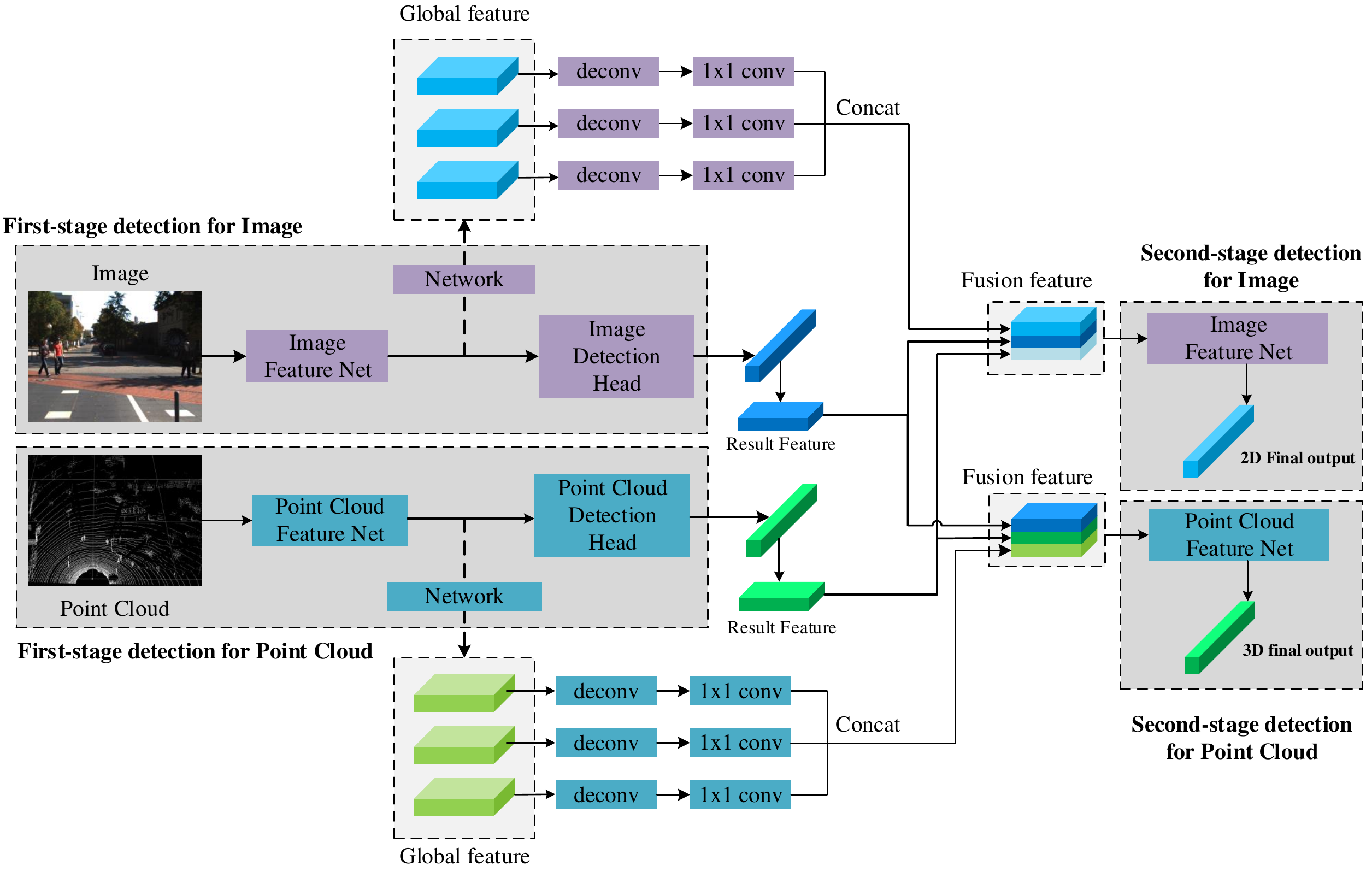}}
\caption{Overview of MMRD proposed in this study. This model employs a result feature level fusion approach for construction. Result features are generated based on the first-stage detection results, which are combined with global features to obtain fusion features. These features can be used for second-stage detection, which can result in more accurate object detection.}
\label{fig1}
\end{figure*}

In this paper, MMDR model is proposed based on result feature-level fusion which is shown in Fig.~\ref{fig1}. The inference process of the multimodal model can be divided into below three stages.

\textit{1)} The first stage of the model consists of two first-stage detectors, which are used to extract preliminary features from the point cloud and image inputs and obtain detection results. Meanwhile, in order to obtain the global and background information of the image and point cloud in the subsequent feature fusion stage, the multi-scale features which are extracted by image feature network or point cloud feature network are preserved.

\textit{2)} The second stage is the feature fusion part. In this stage, the multi-scale features are transformed by deconvolution and pointwise convolution ($1\times1 \ conv$) for scale transformation and integrated into image or point cloud global features.
The first-stage detection results are also transformed into image or point cloud result features in a certain form.
Finally, based on the global features and result features extracted above, fusion features are integrated. The image fusion features consist of image result features, point cloud result features, and image global features, while the point cloud fusion features consist of point cloud result features, image result features, and point cloud global features.

\textit{3)} The third stage consists of two second-stage detectors for further feature extraction of the fusion features and obtaining the final detection results. 
For the fusion features of the image and point cloud, image post-processing feature network and point cloud post-processing feature network are respectively used for further feature calculation and outputting detection results.

Compared with previous decision-level and feature-level multi-modal object detection algorithms, the proposed MMDR model in this study has the following advantages.

\textit{1)} The MMDR model generates features from the first-stage detection results, which can better utilize deep features and reduce the complexity of subsequent feature extraction, enabling the model to focus more on effective information. 

\textit{2)} The MMDR model fuses the global features of the single modalities, which can represent background information, thus alleviating the problem of information loss compared with decision-level fusion.

\textit{3)} The MMDR model has a wider range of applications, can be applied to both 2D and 3D detection tasks, and can handle data from different modalities. 

\subsection{Details of multimodal fusion}
\subsubsection{Extraction of global features}
For obtaining global features, the MMDR model utilizes multi-scale features extracted from the first-stage point cloud or image detection backbone network and processes them accordingly. 
For the image input, the input information has a size of $(640, 640, 3)$, where $640$ represents the pixel size of the input image, and $3$ represents the number of channels indicating a color image. The input image undergoes preprocessing and normalization and features are extracted by the backbone network. 
During the feature extraction process, multi-scale features with dimensions of $(40, 40, 1024)$, $(80, 80, 512)$, and $(160, 160, 256)$ are produced. 
These multi-scales features can represent the depth of feature extraction and the size of the receptive field, thereby expressing detection targets of different sizes. For point cloud features, the feature network can also be used to obtain multi-scale features with different dimensions.

To aggregate the extracted multi-scale features, the deconvolution layer is utilized to convert them into the same size. Then, Pointwise convolutions are applied to transform the number of channels, and the transformed features are merged to obtain the global feature.

\begin{figure}[htbp]
\centerline{\includegraphics[width = 8.5cm]{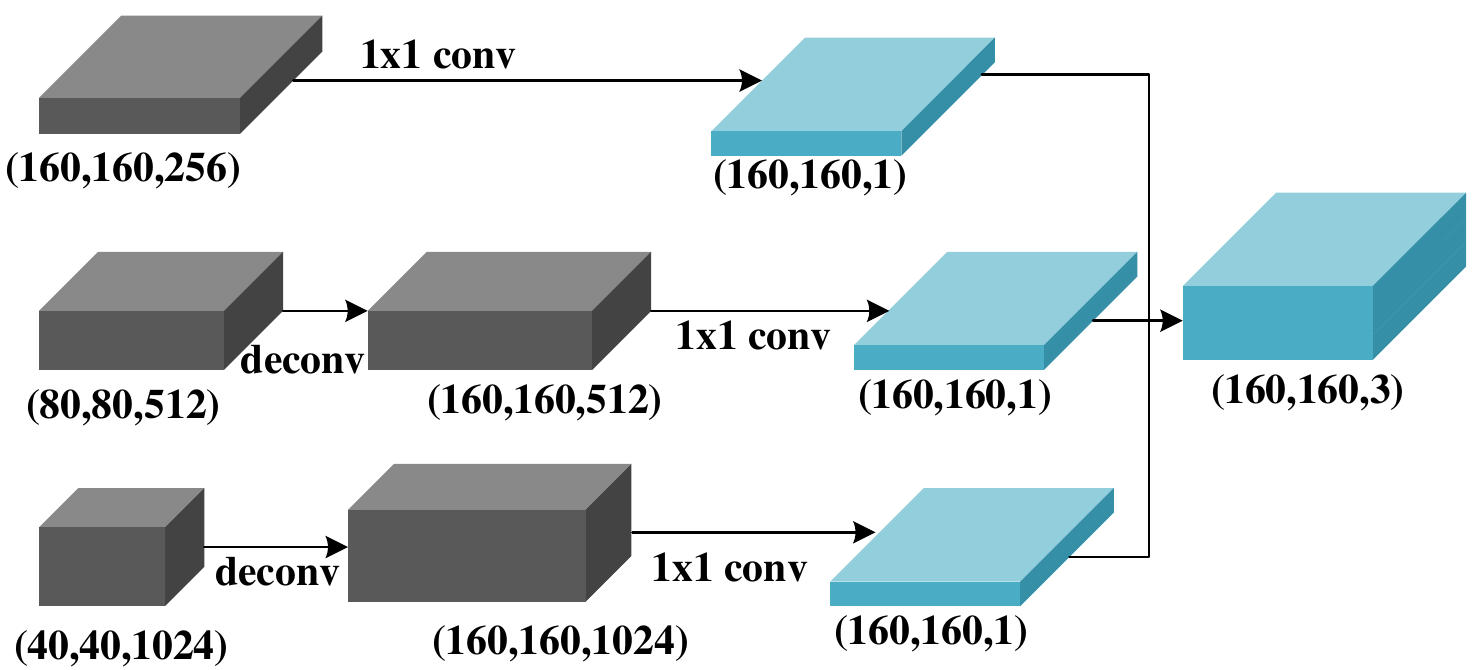}}
\caption{The process of global feature extraction. Pointwise convolution and deconvolution are utilized in this stage to transform the features to corresponding shape.}
\label{fig3}
\end{figure}

After transpose convolution, the feature dimensions can be transformed to $(160, 160, 1024/512/256)$. Then, Pointwise convolution is applied using a kernel size of $(160, 160, 1)$ to transform the features into dimensions of $(160, 160, 1)$. Finally, the output features are concatenated along the channel dimension to obtain the fused feature, with dimensions of $(160, 160, 3)$. The data in different channels represents the whole features at different scales. 
The process of global feature extraction is shown in Fig.~\ref{fig3}.

\subsubsection{Generation of result features}
\begin{figure}[htbp]
\centerline{\includegraphics[width = 8cm]{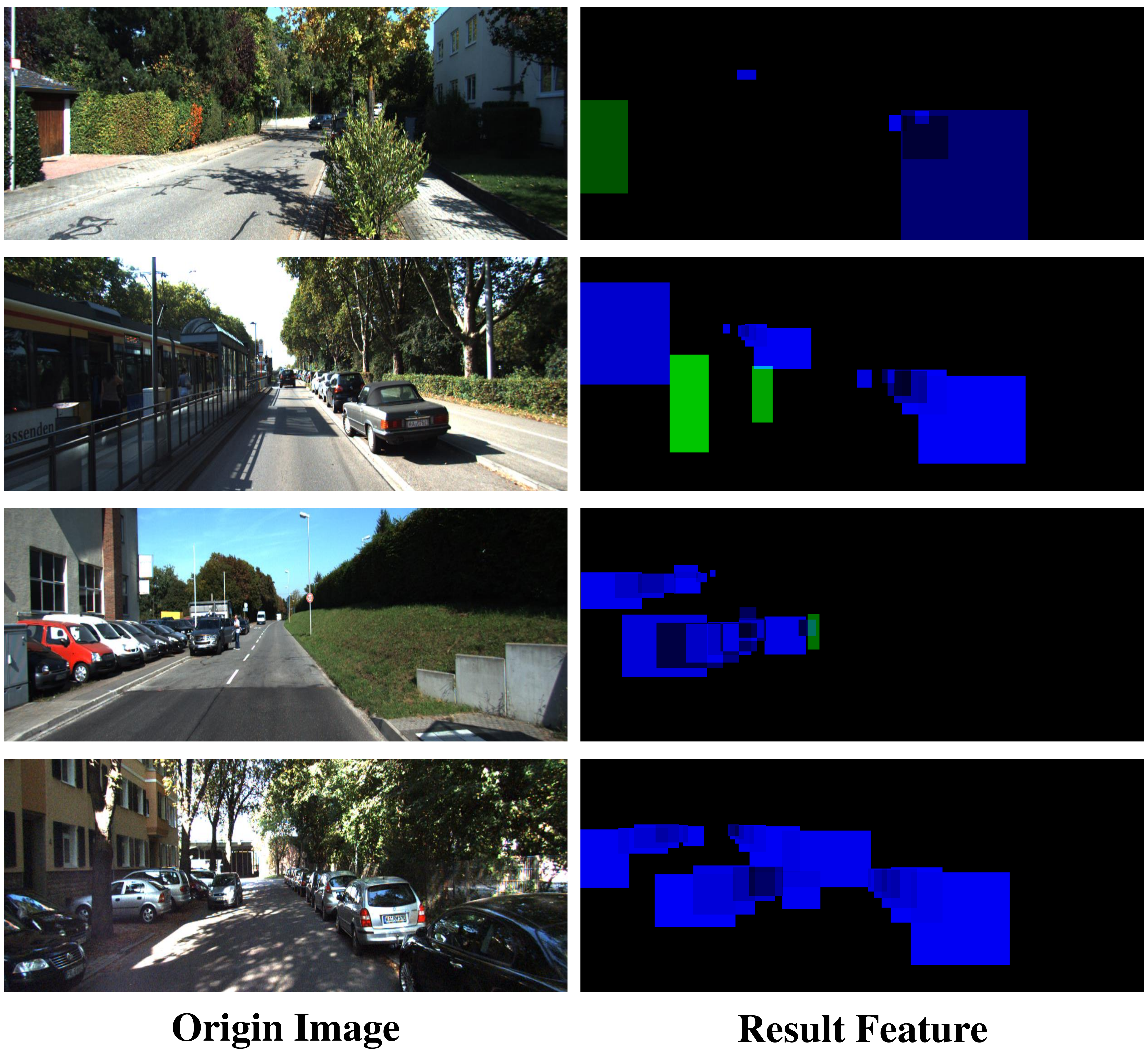}}
\caption{Visualization of the original image and the resulting features.}
\label{rf}
\end{figure}

In this study, the result features are generated from the detection results of the first-stage detector.
For the point cloud/image first-stage detector, the detection results set $\Omega$ consist of  many single detection results and they can be expressed as the format of $(x_{1},y_{1},x_{2},y_{2},s,cls)$. 
Here, $(x_{1,i},y_{1,i},x_{2,i},y_{2,i},s_{i},cls_{i})$ denotes a single detection data in the $i-th$ detection result set. 
$(x_{1,i},y_{1,i},x_{2,i},y_{2,i})$ represents the normalized detection box results, $s$ is the confidence of the detection result, and $cls$ is the category of the detection result, with a value range of $[0,1,2]$, representing pedestrians, vehicles, and cyclist. 
The detection result features can be obtained based on the above results using Algorithm.~\ref{alg:1}.
The visualization results of the features obtained from the algorithm on three channels are shown in Fig.~ \ref{rf}.
 
\begin{algorithm}
    \setlength{\baselineskip}{16bp}
    \caption[]{Generation of result features}
    \label{alg:1}
        \begin{algorithmic}[1]
            \REQUIRE $\Omega$, where $\Omega = \{(x_{1,i},y_{1,i},x_{2,i},y_{2,i},s_{i},cls_{i})\}$
            \STATE Initialize $Opt$ as size $160\times 160$ matrix with $Opt[i,j]=0$ 
            \FOR{$R \in \Omega, R=(x_{1,i},y_{1,i},x_{2,i},y_{2,i},s_{i},cls_{i})$}  
            \FOR{$i \in [R[0],R[2]]$}  
            \FOR{$j \in [R[1],R[3]]$}
            \STATE Update $Opt[i, j, R[5]] = R[4]$ 
            \ENDFOR
            \ENDFOR
            \ENDFOR
        \end{algorithmic}  
\end{algorithm}
 
\subsubsection{Fusion of multiple features}
\begin{figure}[htbp]
\centerline{\includegraphics[width = 7cm]{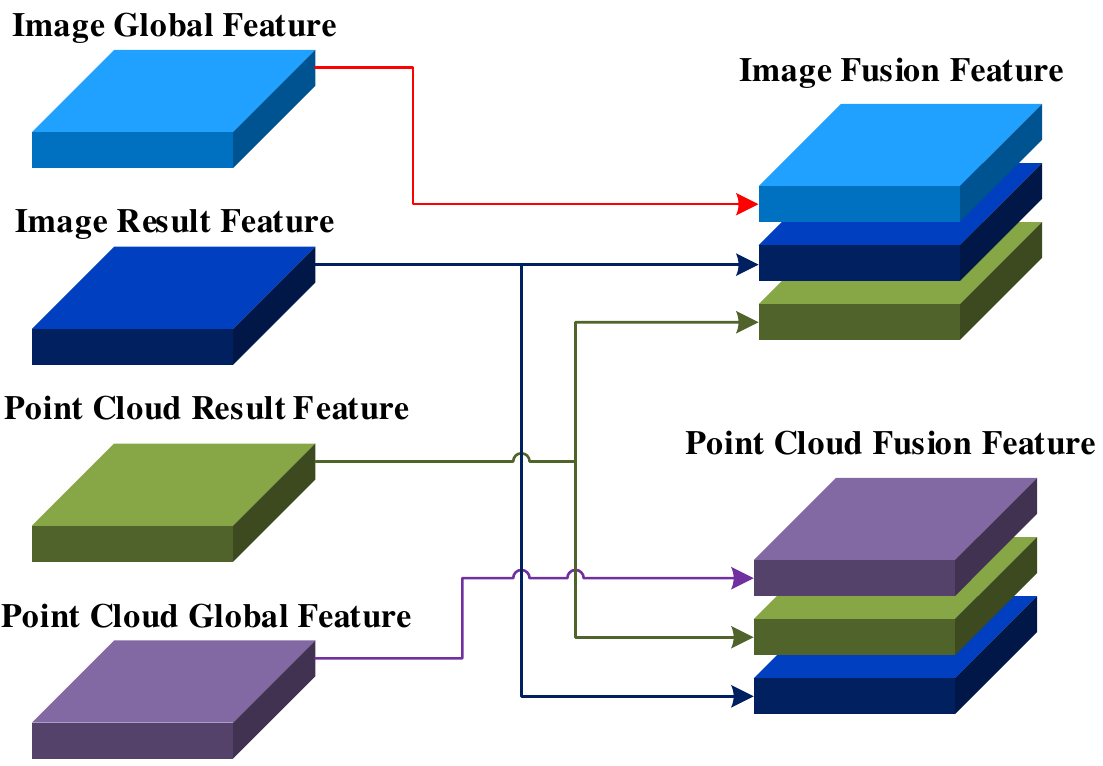}}
\caption{Features are fused to generate fusion features.}
\label{multi_feature_merge}
\end{figure}

In the proposed MMDR model of this paper, some features including point cloud result feature, image result feature, point cloud global feature, and image global feature are fused together to generate the final fusion feature. During the process of feature fusion, the size of each feature is $(160,160,3)$, and they are concatenated as shown in Fig.~\ref{multi_feature_merge}.

For the point cloud fusion feature, a combination of the point cloud and image result feature and the point cloud global feature is used. The result feature is compressed into a two-dimensional form during the feature extraction process. In the point cloud detection task, the purpose of result feature lies in providing the feature of interest area under both 2D and 3D views, facilitating the further feature extraction and 3D detection box prediction based on the global feature.

For the image fusion feature, a combination of the point cloud and image result feature and the image global feature is used. At this stage, the result feature provides the main support for the final 2D detection result prediction. Meanwhile, the image global feature provides the background and global feature, alleviating the problem of missed detection.

\subsubsection{second-stage detection}
\begin{figure}[htbp]
\centerline{\includegraphics[width = 8.5cm]{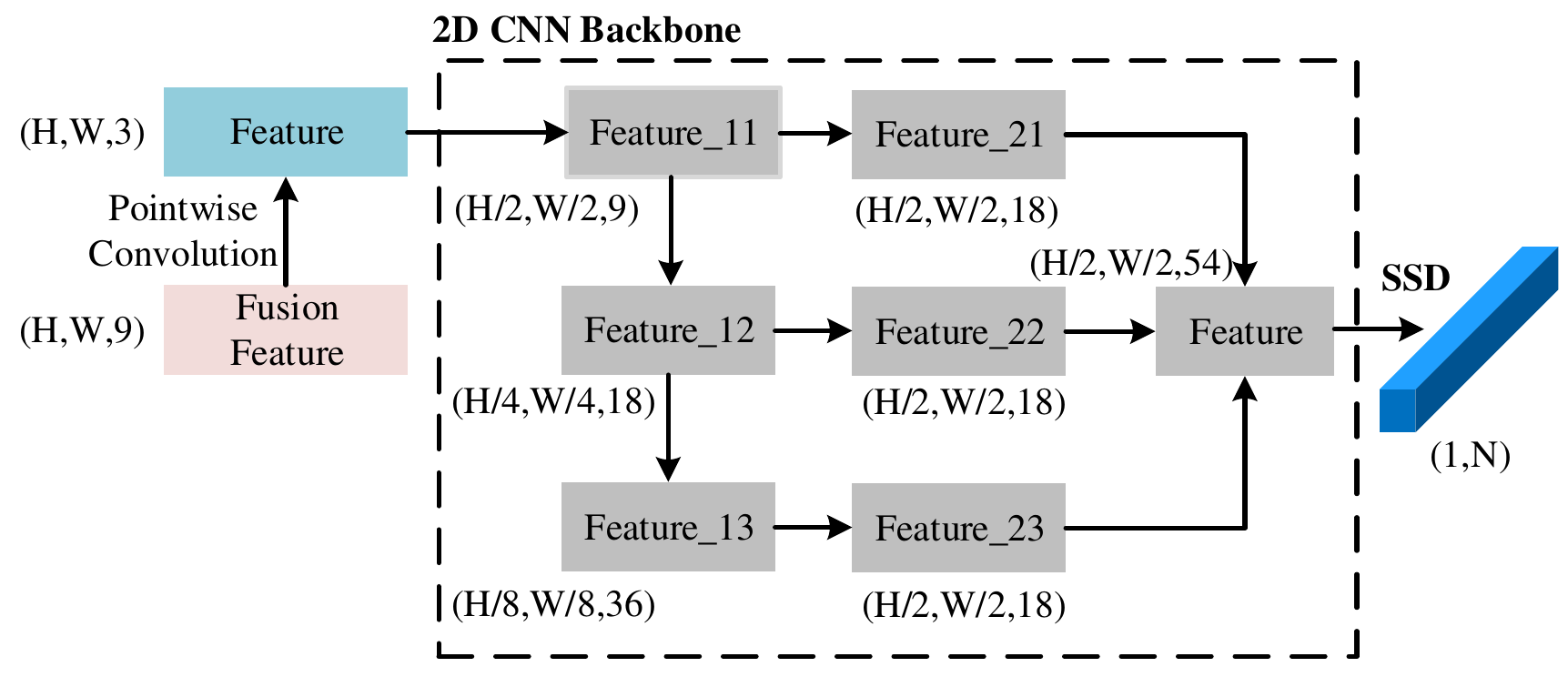}}
\caption{Multiple feature fusion to generate fusion features.}
\label{fig:mmdr_3h}
\end{figure}

For the second-stage detection of point clouds, the MMDR model uses a backbone and detection head similar to single modality object detection to process the fused features and output the corresponding detection results, as shown in Fig.~\ref{fig:mmdr_3h}.

For the second-stage detection on images, the backbone and detect head of YOLOX\cite{ge2021yolox} were referenced for the design. Considering that the input of the model in the second-stage detection contains fused point cloud results, the parts of the image that are not visible in the field of view or are occluded are visible in the perspective of the feature extraction network. This performance will actually have an impact on downstream applications of 2D object detection results.

\begin{figure}[htbp]
\centerline{\includegraphics[width = 7cm]{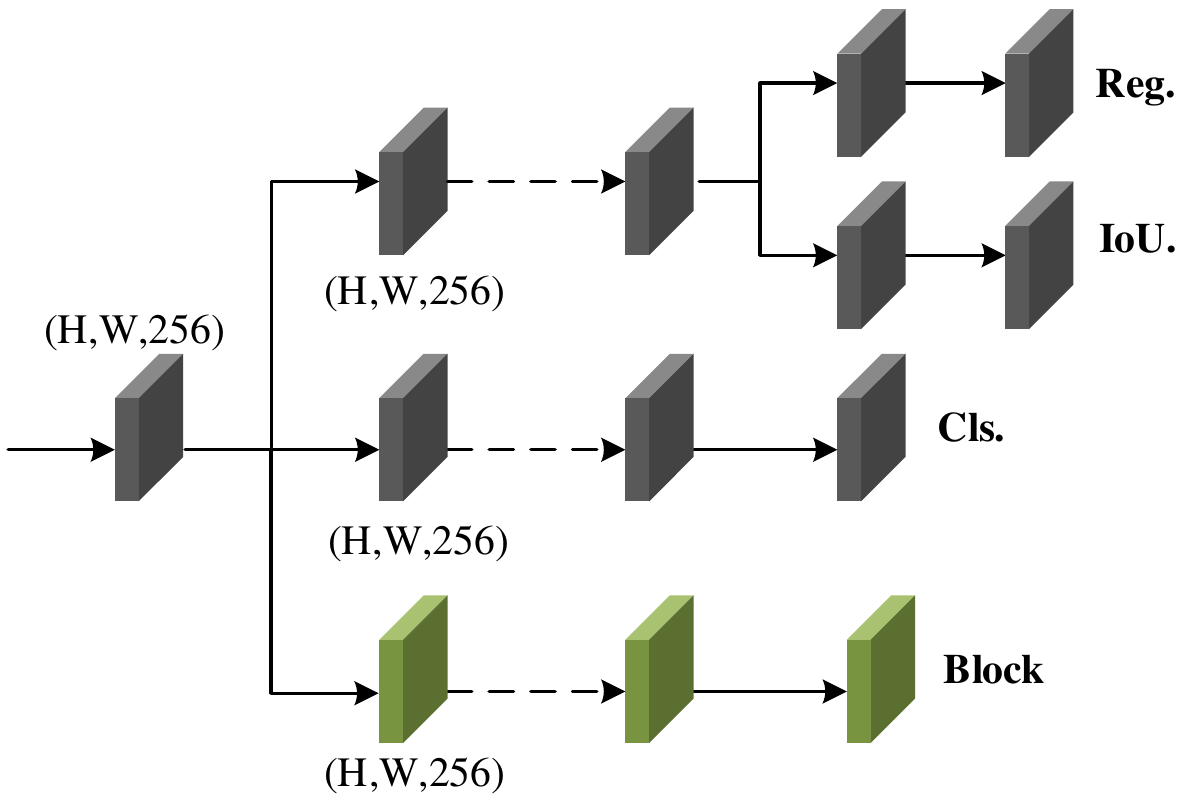}}
\caption{The image second-stage detection decoupled head in MMDR model.}
\label{fig:mmdr_2h}
\end{figure}

Therefore, in the output stage of the 2D detection results, the MMDR model added an extra prediction for whether there is occlusion (block), unlike the decoupled head of YOLOX. The image second-stage detection head of MMDR is shown in Fig.~\ref{fig:mmdr_2h}. The variable for whether there is occlusion (block) is a binary variable, with a true value of 0 (no occlusion) or 1 (occlusion).

The loss function of the image second-stage detection can be expressed as shown in Eq.~\ref{eq63}. Here, $L_{obj}, L_{cls}, L_{reg}$ are computed using the same method as YOLOX, while $L_{block}$ is computed according to Eq.~\ref{eq64}. Here, $\hat{y}$ represents the predicted result of whether there is occlusion in the model, and $y$ represents the ground truth label.
\begin{equation}
\label{eq63}
L_{2D-2nd} = \beta_{1}L_{cls} + \beta_{2}L_{reg} + \beta_{3}L_{obj} + \beta_{4}L_{block}
\end{equation}
\begin{equation}
\label{eq64}
L_{block}=-(y\cdot \log(\hat{y})+(1-y)\cdot \log(1-\hat{y}))
\end{equation}

\section{Experiments}

\subsection{Dataset and platform}
The experiments for the proposed MMDR model were conducted on the KITTI detection dataset \cite{kitti} to evaluate its performance against baseline 2D and 3D detection models.
The KITTI provides 7481 labeled RGB images and Velodyne point clouds, and 7518 RGB images and Velodyne point clouds for test. Also, there are 20 groups of tracking data for training and 28 groups of data for test. RGB images, Velodyne point clouds, camera calibration matrices, and others are included in this data set. 

\begin{figure}[htbp]
\centerline{\includegraphics[width = 7cm]{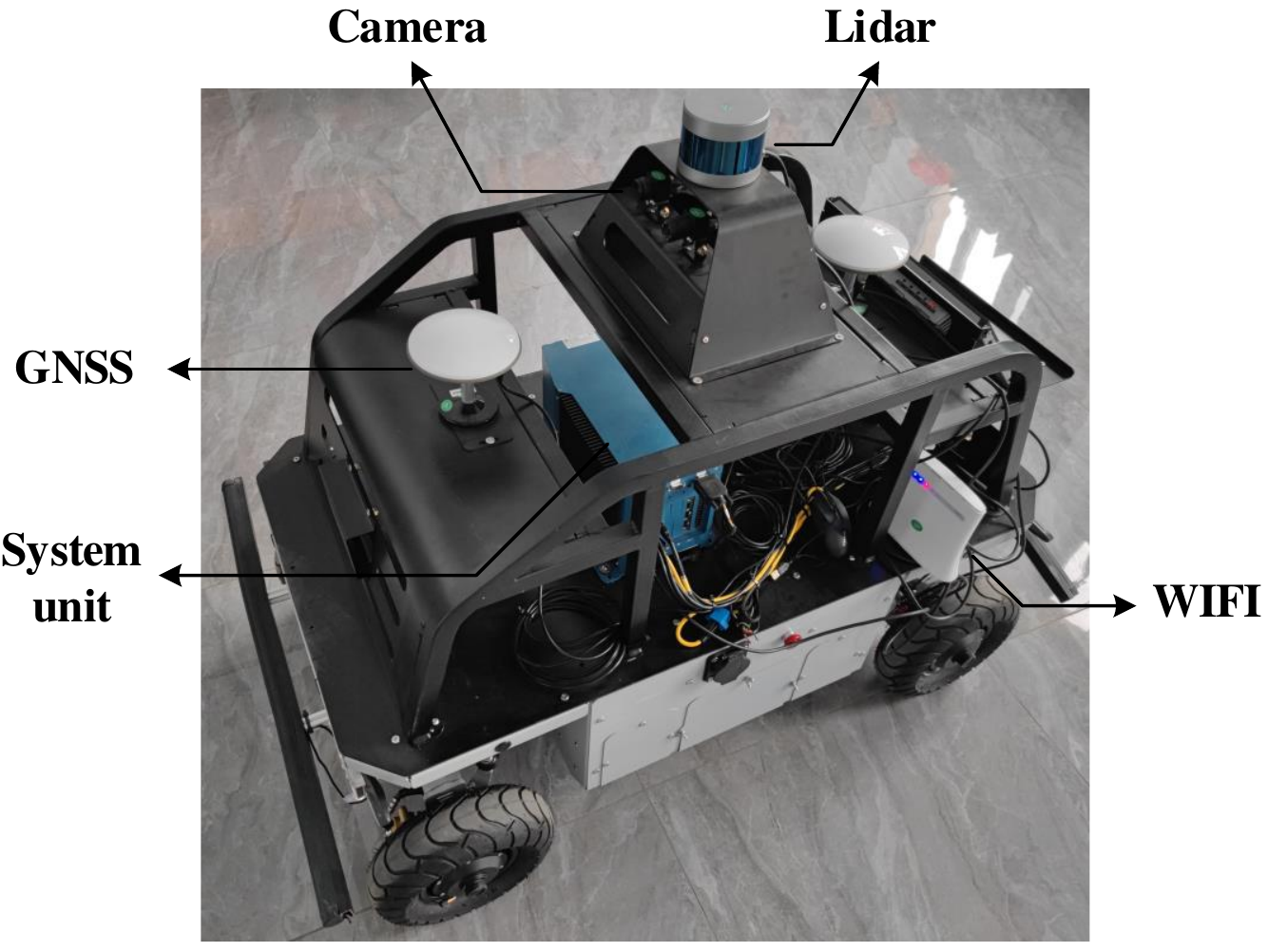}}
\caption{The Apollo autonomous platform which is utilized to conduct experiments in real environments.}
\label{apollo}
\end{figure}

Additionally, this study conducted information acquisition and simulation verification in real-world campus environments based on Baidu's Apollo autonomous driving platform which is shown in Fig.~\ref{apollo}. 
Apollo autonomous platform is equipped with multiple sensors and data processing modules, such as LiDAR, camera, GPS, GPU, etc., to collect various information of the real environment.

\begin{figure*}[htbp]
\centerline{\includegraphics[width = 17.5cm]{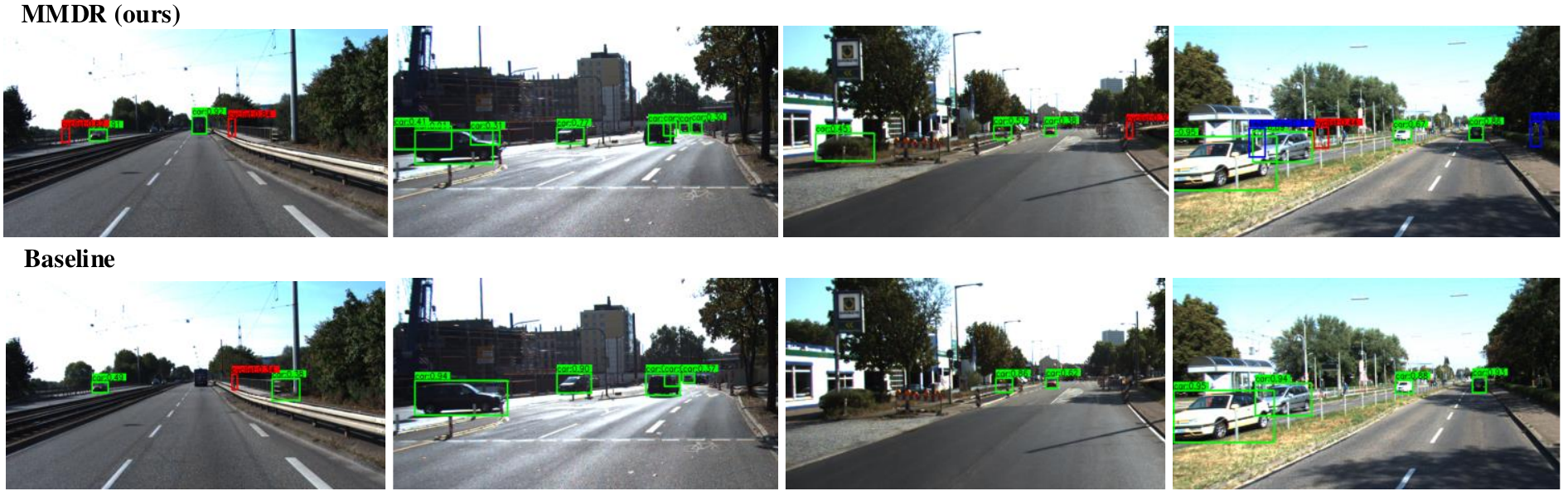}}
\caption{The visualization of the comparison results for 2D object detection. The images above show the 2D detection results of the baseline model on images, while the images below show the detection results of the proposed MMDR model on the same dataset.}
\label{2dres}
\end{figure*}

\begin{figure*}[htbp]
\centerline{\includegraphics[width = 17.5cm]{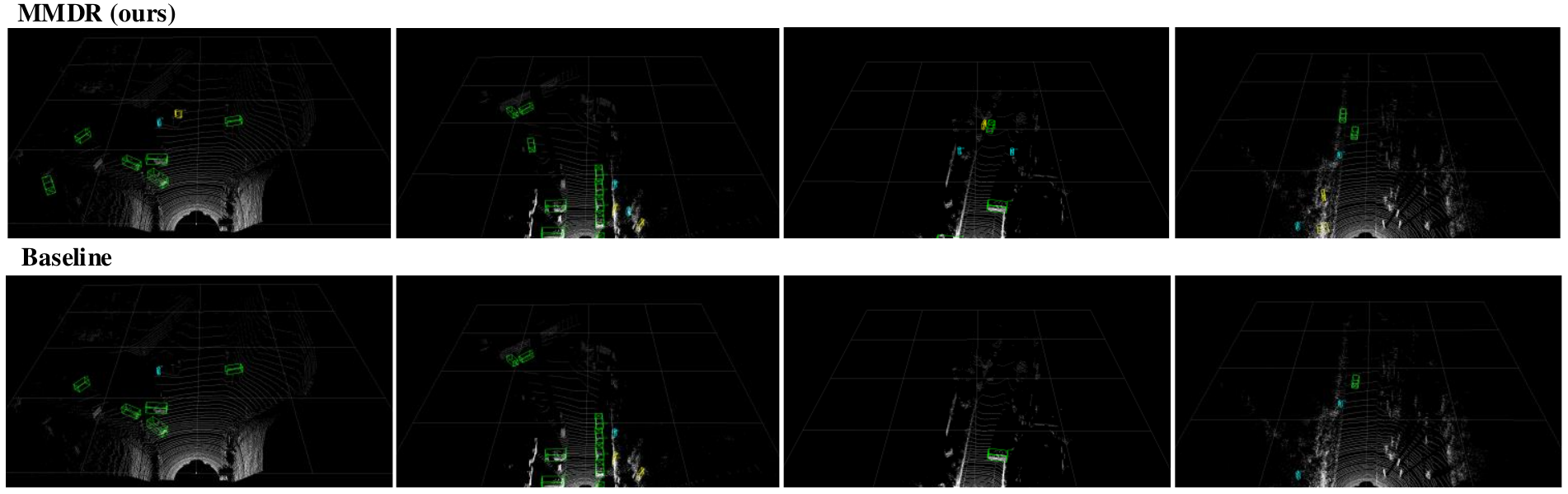}}
\caption{The visualization of the comparison results for 3D object detection on point clouds from BEV. The images above show the 3D detection results of the baseline model on point cloud, while the images below show the detection results of the proposed MMDR model on the same dataset.}
\label{3dres}
\end{figure*}

\subsection{2D Detection on KITTI}
For the 2D detection task, the performance of MMDR's 2D detection output was compared with the baseline detector YOLOX. The performance metrics used were Average Precision (AP) and Average Recall (AR). The experimental results are shown in Table.~\ref{table1}. In addition, this paper also compared the performance of the MMDR model with other classical image object detection models in terms of 2D detection of pedestrians and vehicles, which is shown in Table.~\ref{table3}. 
The results of the comparison demonstrate that the MMDR model outperforms the other models in terms of accuracy and robustness. This highlights the effectiveness of the proposed approach for improving the detection of important objects in autonomous systems.

\begin{table}[ht]
    \centering
    \renewcommand\arraystretch{1.6}
	\caption{Comparison of 2D detection on KITTI with baseline.}
	\setlength{\tabcolsep}{2mm}{
	\label{table1}       % Give a unique label
    \begin{tabular}{cccccl}
    \hline
    \multirow{2}{*}{\textbf{Metric}} & \multirow{2}{*}{\textbf{Model}} & \multicolumn{4}{c}{\textbf{KITTI class}}                                                 \\

                            &                        & \textbf{Car}   & \multicolumn{1}{l}{\textbf{Pedestrian}} & \multicolumn{1}{l}{\textbf{Cyclist}} & \textbf{All}  \\
                                \hline 
    \multirow{2}{*}{AP}     & YOLOX                  & 80.16 & 53.75                          & 54.82                       & 62.9 \\
                            & MMDR(ours)             & 85.49 & 60.70                          & 69.23                       & 69.2 \\
                            \hline
    \multirow{2}{*}{AR}     & YOLOX                  & 84.04 & 61.61                          & 62.67                       & 71.8 \\
                            & MMDR(ours)             & 88.64 & 68.58.                         & 75.14                       & 76.7 \\
    \hline 
    \end{tabular}
	}
\end{table}

\begin{table}[ht]
    \centering
    \renewcommand\arraystretch{1.6}
	\caption{Comparison of 2D detection on KITTI with classic detection methods.}
	\setlength{\tabcolsep}{0.4mm}{
	\label{table3}       % Give a unique label
    \begin{tabular}{cccc}
    \hline
    \multirow{2}{*}{\textbf{Model}} & \multirow{2}{*}{\textbf{Modal}} & \multicolumn{2}{c}{\textbf{KITTI class}}                                                      \\
                                   &                                 & \textbf{Car} & \multicolumn{1}{l}{\textbf{Pedestrian}}  \\ \hline
		Fast R-CNN ResNet50                & Image     & 73.6 & 58                                                  \\
		Fast R-CNN ResNet101                & Image      & 72.6 & 57.9                                               \\
        Fast R-CNN Inception\_ResNet\_V2                & Image      & 72.1 & 57.2                                             \\
		Fast R-CNN Inception\_V2         & Image      & 74.5 & 53                                                 \\
		SSD Inception\_V2            & Image               & 47.8 & 27.9                                               \\
		SSD MobileNet\_V2             & Image               & 46.2 & 27.5                                               \\
		SSD MobileNet\_V1        & Image               & 42.7 & 25.2                                             \\
		\hline
		MMDR                & Lidar \& Image      & 85.49 & 60.70                               \\
    \hline
    \end{tabular}
	}
\end{table}

The results are visualized in Fig.~\ref{2dres}. The comparison results clearly indicate that the MMDR model proposed in this study outperforms the original baseline model in terms of detection performance. This is due to the fact that the MMDR model fuses 3D feature information, enabling it to detect more comprehensive targets with better accuracy.

\subsection{3D Detection on KITTI}
For 3D detection tasks, some experiments are conducted on the MMDR model and the PointPillars baseline network in our study, using the KITTI dataset. We also studied the AP performance of the related models in 2D detection boxes from both a 3D detection perspective and a bird's-eye view (BEV) perspective in the KITTI dataset. The specific experimental results are shown in Table.~\ref{table2}.

Moreover, MMDR model proposed in this paper with other classic 3D detection algorithms is compared on its detection accuracy. The results are shown in Table.~\ref{table4}. 

\begin{table}[ht]
    \centering
    \renewcommand\arraystretch{1.6}
	\caption{Comparison of 3D detection on KITTI with baseline.}
	\setlength{\tabcolsep}{2mm}{
	\label{table2}       % Give a unique label
    \begin{tabular}{ccccc}
    \hline
    \multirow{2}{*}{\textbf{View}} & \multirow{2}{*}{\textbf{Model}} & \multicolumn{3}{c}{\textbf{KITTI class}}                                                      \\
                                   &                                 & \textbf{Car} & \multicolumn{1}{l}{\textbf{Pedestrian}} & \multicolumn{1}{l}{\textbf{Cyclist}} \\ \hline
    \multirow{2}{*}{3D}            & PointPillars                    & 69.55        & 58.44                                   & 54.68                                \\
                                   & MMDR                     & 70.67        & 59.17                                   & 56.02                                \\
                                   \hline
    \multirow{2}{*}{BEV}           & PointPillars                           & 75.98        & 65.46                                   & 62.14                                \\
                                   & MMDR                   & 76.65        & 66.42                                   & 62.76                                \\ \hline
    \end{tabular}
	}
\end{table}

\begin{table}[ht]
    \centering
    \renewcommand\arraystretch{1.6}
	\caption{Comparison of 3D detection on KITTI with classic detection methods.}
	\setlength{\tabcolsep}{2mm}{
	\label{table4}       % Give a unique label
    \begin{tabular}{ccccc}
    \hline
    \multirow{2}{*}{\textbf{Model}} & \multirow{2}{*}{\textbf{Modal}} & \multicolumn{3}{c}{\textbf{KITTI class}}                                                      \\
                                   &                                 & \textbf{Car} & \multicolumn{1}{l}{\textbf{Pedestrian}} & \multicolumn{1}{l}{\textbf{Cyclist}} \\ \hline
		MV3D                & Lidar \& Image      & 55.12 & N/A                            & N/A                         \\
		AVOD                & Lidar \& Image      & 66.38 & 50.80                          & 46.61                       \\
		F-PointNet          & Lidar \& Image      & 62.19 & 51.21                          & 50.39                       \\
		VoxelNet            & Lidar               & 57.73 & 39.48                          & 48.36                       \\
		SECOND              & Lidar               & 66.20 & 51.07                          & 53.83                       \\
		PointPillars        & Lidar               & 69.55 & 58.44                          & 54.68                       \\
		\hline
		MMDR                & Lidar \& Image      & 70.67 & 59.17                          & 56.02     \\
    \hline
    \end{tabular}
	}
\end{table}

By analyzing the results of Table.~\ref{table2} and Table.~\ref{table3}, the effectiveness of the proposed model has been validated through a series of rigorous experiments and comparative analyses with other models. 

It can be concluded that the multimodal MMDR proposed in this paper has demonstrated significant improvement in 3D detection performance when compared to classic models. The combination of different types of sensor data and the use of a novel multimodal fusion strategy have contributed to the increased accuracy the proposed model.
\begin{figure*}[htbp]
\centerline{\includegraphics[width = 17.5cm]{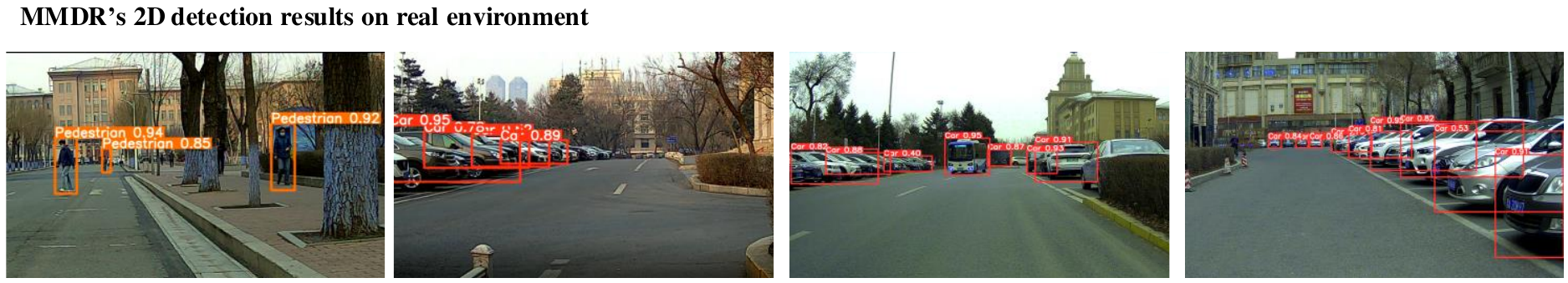}}
\caption{MMDR's 2D detection results on real campus environment based on images and point clouds.}
\label{apollo_2d}
\end{figure*}
\begin{figure*}[htbp]
\centerline{\includegraphics[width = 17.5cm]{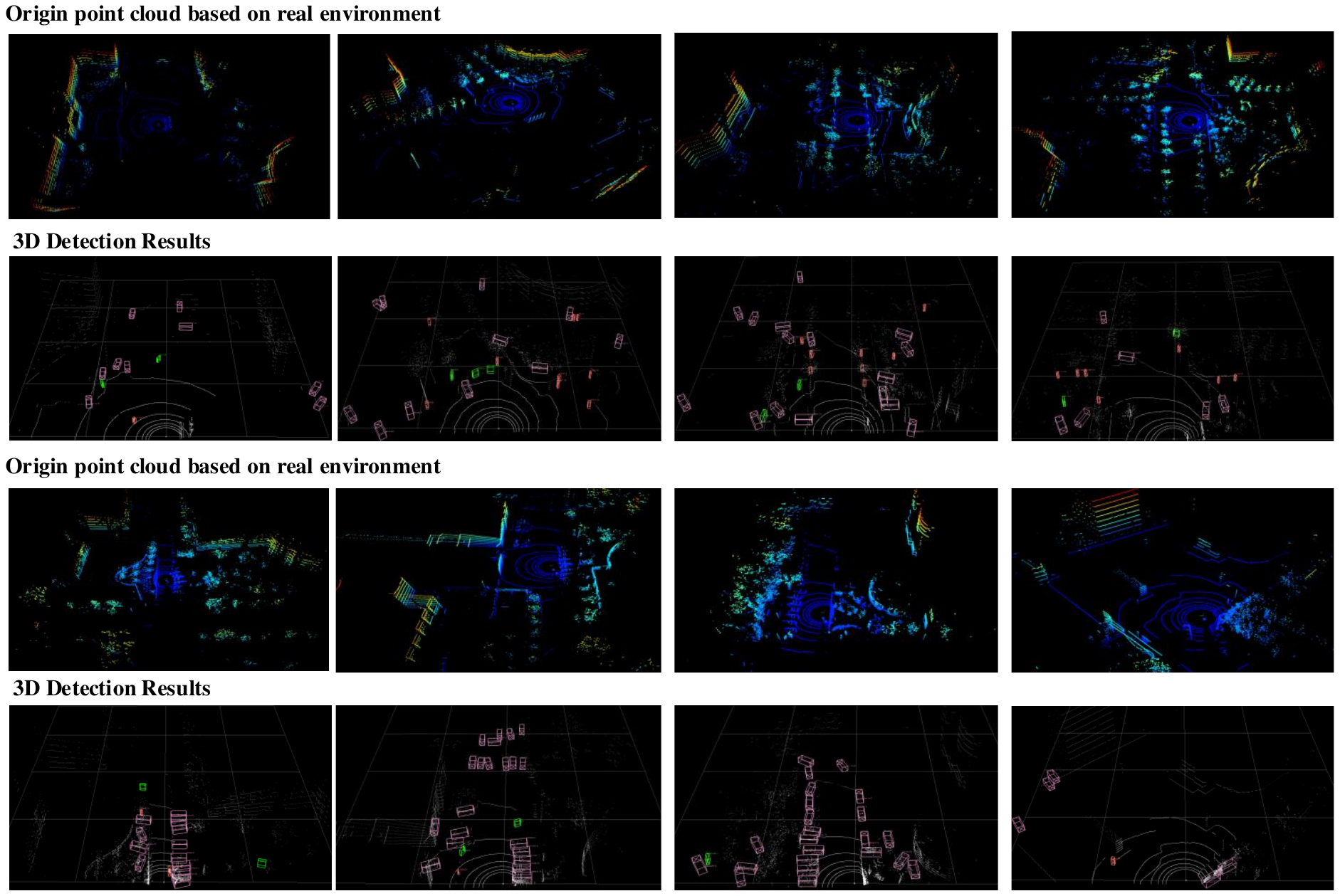}}
\caption{MMDR's 3D detection results on a real campus environment which based on both images and point clouds. The visualization of the original point clouds and the detection results on the 3D view are presented.}
\label{apollo_3d}
\end{figure*}

\subsection{Experiments on Apollo}
To further validate the feasibility of the algorithm proposed in this paper in practical physical environments, data collection and experimental verification were conducted on the Apollo experimental platform. The experimental data was collected from the actual campus environment.

Fig.~\ref{apollo_2d} illustrates the 2D detection results of the model in the actual environment.
Fig.~\ref{apollo_3d} presents some examples of the model's 3D detection results in the actual environment.
The mentioned results demonstrate the feasibility of the algorithm in real-world environments.

\section{Conclusion}
This paper proposes a multimodal fusion approach based on result feature-level fusion, which utilizes the result features generated from single modality sources and fuses them for downstream tasks. This method fully utilizes the detection result message of a single modality. 
Based on this method, the MMDR model has been proposed that enables better representation of the deep-level features of single modality sources and incorporates the global feature of a single modality. 
The MMDR model can take into account both the result features of a single modality and the background information simultaneously, thus avoiding issues such as detection loss while improving detection accuracy.
We conducted experiments on the KITTI benchmark for 2D and 3D detection tasks, and the results show that the MMDR model outperforms the baseline algorithms. Moreover, we validated the proposed model in a real campus environment using the Apollo autonomous system, and the experimental results confirm its feasibility in a real environment.
It is worth noting that the proposed approach does not fully utilize all messages recorded from other sensors. Thus, a potential area for improvement in our future work is to fuse multiple sensor information to further improve the performance in a real environment.

\bibliographystyle{IEEEtran}
\bibliography{reference.bib}

\end{document}